# What can we learn from slow self-avoiding adaptive walks by an infinite radius search algorithm?

Susan Khor

**Abstract**

Slow self-avoiding adaptive walks by an infinite radius search algorithm (*Limax*) are analyzed as themselves, and as the network they form. The study is conducted on several NK problems and two HIFF problems. We find that examination of such "slacker" walks and networks can indicate relative search difficulty within a family of problems, help identify potential local optima, and detect presence of structure in fitness landscapes. Hierarchical walks are used to differentiate rugged landscapes which are hierarchical (e.g. HIFF) from those which are anarchic (e.g. NK). The notion of node viscosity as a measure of local optimum potential is introduced and found quite successful although more work needs to be done to improve its accuracy on problems with larger K.

## 1. Introduction

Knowing the essential features of a problem's fitness landscape helps in the design and tuning of heuristic search algorithms to increase their chance of success. However, this logic is recursive since differently designed search algorithms, or even same search algorithms differently tuned, can produce different fitness landscapes for the same problem. The problem lies in the widely accepted definition of a search point's neighbourhood in a fitness landscape [e.g. Jones 1995; Ochoa et al 2008; Verel et al 2008]. To somewhat circumvent this measuring ruler problem, we use a search algorithm with an infinite radius (*Limax*) to saunter the search space of a problem.

*Limax* has the following pivot rule: move to a not already visited nearest neighbour (this is the slow part of a walk[1]) solution which gives a fitness improvement over the current solution. Since *Limax* has infinite radius - it can reach any solution from any other solution in a search space - all its self-avoiding adaptive walks terminate at a global optimum. All such walks (one is initiated from every solution) and the step sizes (distances between solutions in terms of Hamming distance) they take are examined in the following two ways.

---

[1] "The shortest distance between two points is often unbearable." – Charles Bukowski

First, *Limax* walks are analyzed in terms of distance traveled, step size variability, step sequence compressibility and step sequence pattern (section 3). Our hypothesis is that more difficult search problems will yield on average walks which cover farther distances, use a wider range of step sizes, are less compressible, and make no distinctive pattern of steps. Walks which cover farther distance incurs higher risk of going astray and not reaching a global optimum. It also means that on average, solutions are far away from a global optimum and if this distance scales unfavourably with increases in problem size, having a 'funnel' shaped fitness landscape does not guarantee an easy search [Doye, 2002]. A more varied step size implies a larger set of move operations for a stochastic search algorithm and thus more uncertainty as to the right move operation to make at a given time. A less compressible sequence of step sizes implies more frequent changes to the move operation. A step sequence with no discernable step pattern provides little guidance about how to change a move operation and thus increases uncertainty of search success. Taken together, the correct generation of move sequences with more variability, less compressibility and no history to infer from, requires a more complicated stochastic search algorithm that can "come to know" which move operator to use and when correctly. It is in this sense that search space analysis by *Limax* can be related directly to the design of a stochastic search algorithm[2]. Our study confirms a positive relationship between problem difficulty and stochastic search algorithm complicatedness as defined in this paragraph.

Second, a directed weighted network is constructed from all walks and their step data, and network analysis is performed on the resulting *Limax* network (section 4). Note that *Limax* networks are distinct from *Local Optima Networks* (LONs) [Ochoa et al 2008; Verel et al 2008] in that their nodes and edges carry different meanings. Although it may be possible to transform *Limax* networks to create LON-like networks, and to use results from *Limax* network analysis to construct LONs[3]. Nevertheless, our main goal is to explore other ideas in network analysis of search spaces, not new ways of creating LONs. Previously, [Doye and Massen 2005] conducted a network analysis of the potential energy landscape of atomic clusters.

We find *node viscidity* in *Limax* networks to be a rather good indicator of a node's local optimum potential, i.e. how likely the solution represented by the node is a local optimum for a finite radius search algorithm. The ability to locate local optima and/or rugged areas in a search space easily is essential to other search space analysis methods such as reverse hill climbing [Jones 1995, p.96] and even to build LONs, and this ability becomes imperative with larger search spaces. Further network analysis on node viscidity supports our hypothesis of a positive relationship between node viscidity and local optimum potential. Nodes with high viscosity tend to be centrally located in *Limax* networks, reflecting the nature of local optima to be points of attraction in a fitness landscape. Examination of node viscosity mixing reveals a negative relationship between disassortativity and search difficulty within the NK problems. Given N, as K increases, node viscosity mixing becomes less negatively correlated, indicating a rougher fitness landscape as nodes with higher local optimum potential get closer to each other in a *Limax* network (but not necessarily in terms of Hamming distance which is the measure for the "Massive Central" phenomenon). Not surprisingly, the HIFF problems have negligible node viscosity correlation with assortativity coefficient [Newman 2002] around 0.0. Finally, the "Massive Central" phenomenon [Kauffman 1993, p. 60] described of NK problems could be detected amongst nodes with high viscosity within the NK problems (section 4).

---

[2] Although we do not rule out the possibility that *Limax* itself can be utilized as a stochastic search algorithm.
[3] For example, using nodes with high viscidity to select potential local optima and then reverse hill climbing from them.

**2. Materials**

*2.1 Test Problems*

The basic set of test problems is the NK problems with random neighbourhood interactions [Kauffman 1993]. We use N=14 with K = 2, 6 and 10; and N=16 with K= 4, 8 and 12. A binary alphabet is used giving a search space of $2^N$ points. Since NK problems rely on random values for fitness evaluation, we generated 30 independent instances (both neighbourhood and fitness values were randomized) for each NK problem. NK problems have normal fitness distributions. The globally optimal or maximally fit search point for an NK problem is unique. NK problems with larger K values are known to be more difficult in terms of locating the global optimum [Kauffman 1993]. Additional test problems are: OneMax, HIFF-C [Watson 2002, p.121] and HIFF-M [Khor 2009]. These problems are more structured and deterministic than the NK problems (i.e. their inter-variable dependencies and fitness values are not assigned at random), and are used to demonstrate certain points (e.g. section 3.3).

*2.2 Search algorithm: Limax*

Walks and steps data are gathered by *Limax* by starting a self-avoiding adaptive walk from every point in the search space (our method is enumerative at present). *Limax* moves to a not previously visited nearest fitter neighbour solution. At every search point, *Limax* always attempts the smallest move possible first and gradually increases its step size until a solution fitter than the current one is found. Distance between solutions is measured in terms of Hamming distance. There are no limits to the size of the move or step size that *Limax* can make. Therefore, all *Limax* walks terminate at a global optimum. It is easy to envision *Limax*-Δ, where the maximum step size is restricted to Δ. *Limax*-Δ can be used to study how *Limax* networks (section 4) change with Δ.

## 3. Step Analysis

*3.1 Walk length, distance, compressibility and variability*

Walk length (*wlen*) is the number of steps taken in a walk. Compressed walk length (*cwlen*) is the number of steps in a walk whose steps have been compressed (ala Kolmogorov) as follows: replace consecutive steps of the same size with a single step of the size. For example, a walk *w* with steps ⟨1, 1, 2, 3, 2, 2, 2, 5⟩$_w$ is compressed to ⟨1, 2, 3, 2, 5⟩$_{cw}$. Walk distance (*wdist*) is the sum of step sizes taken in a walk. Compressed walk distance (*cwdist*) is the sum of step sizes in a compressed walk. Compression ratio *cr1* measures compressibility of walks in terms of steps, and is *cwlen* / *wlen*. Compression ratio *cr2* measures compressibility of walks in terms of distance, and is *cwdist* / *wdist*. Step size variability (*wvar*) is the number of unique step sizes taken in a walk. To illustrate, the walk *w* in the previous example has *wlen* = 8, *cwlen* = 5, *cr1* = 5/8, *wdist* = 18, *cwdist* = 13, *cr2* = 13/18, and *wvar* = 4. *cr2* was introduced as a consequence of *wdist* being a better measure of problem search difficulty than *wlen* (*wlen* cannot distinguish between walks of the same length but with vastly different step sizes e.g. ⟨1, 1, 2, 3, 2⟩$_w$ and ⟨1, 2, 2, 5, 8⟩$_w$ . Compare Fig. 1 top and middle). Our results confirm that for our purpose, *cr2* does not yield different information from *cr1* (Fig. 1 bottom).

Analysis of walks and steps for each NK problem instance are summarized, and in Figs. 1 and 2 these statistics are averaged over the 30 independent instances for each NK problem. Corresponding statistics are given for the One-Max problem when N=14 to illustrate an easy search problem. Our hypothesis is that more difficult search problems will yield on average walks which cover farther distances, use a larger set of step sizes and are less compressible. Respectively, this translates to larger values for *wdist*, *wvar* and *cr1* as search difficulty increases. And as Figs. 1 and 2 show, this is indeed the case. Given N, NK problems with larger K have significantly longer, varied and less compressible walks.

*3.2 Adaptive length*

Compressing steps also reveals the adaptive length, i.e. the longest sequence of same sized steps in an adaptive walk. Fig. 3 shows the maximum compressed sequence length averaged over 30 instances for each NK problem. Given N, the adaptive length decreases significantly as K increases. This follows from the observation in section 3.1 that the walks of more difficult NK problems are less compressible. From the perspective of a stochastic search algorithm, shorter adaptive lengths imply more frequent changes in step size or the move operator, and hence increased search algorithm complicatedness.

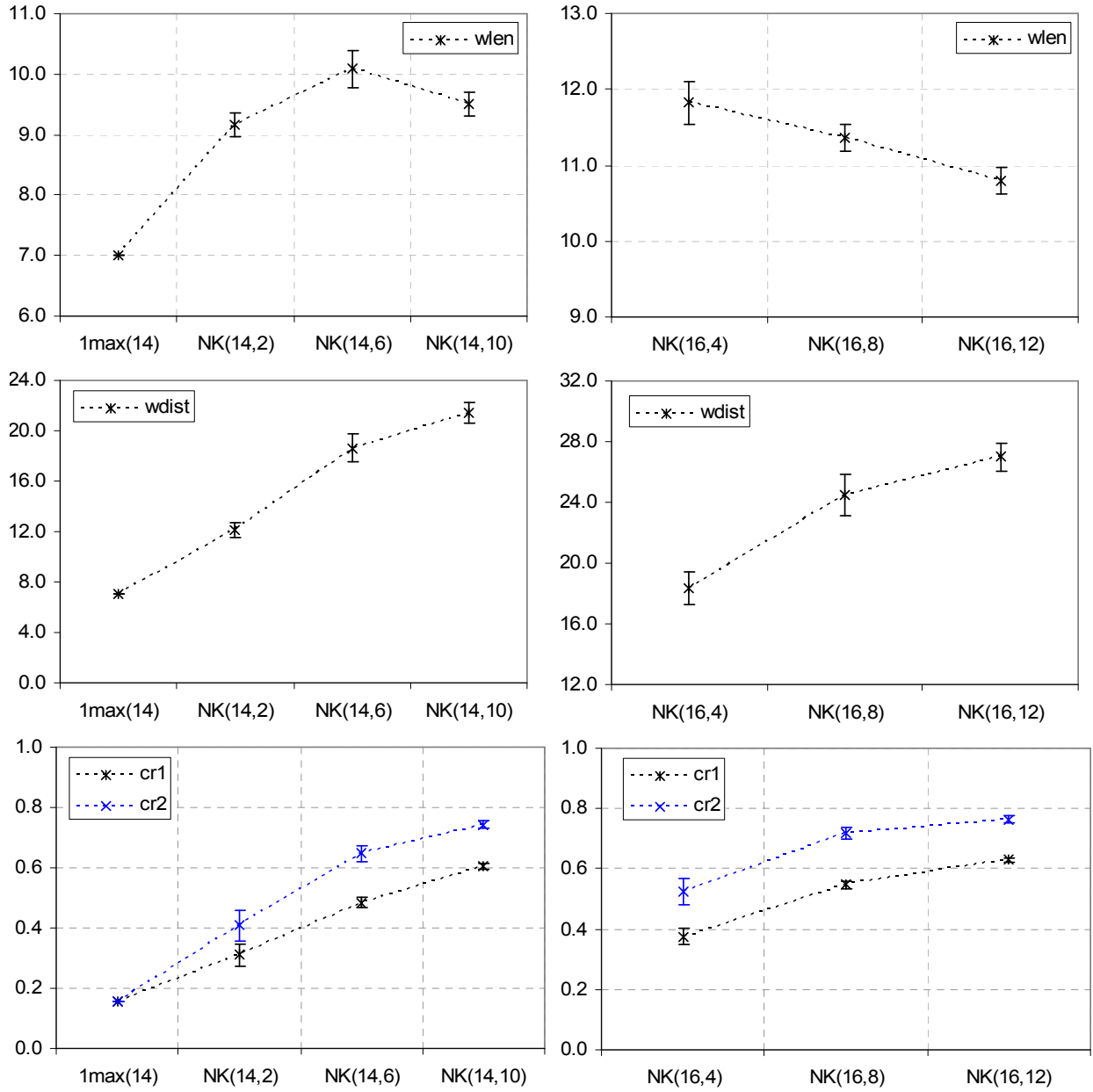

**Fig. 1** Top to bottom: walk length, walk distance and compression ratio averaged over 30 instances per NK problem. See text for explanation. Error bars indicate 95% confidence interval.

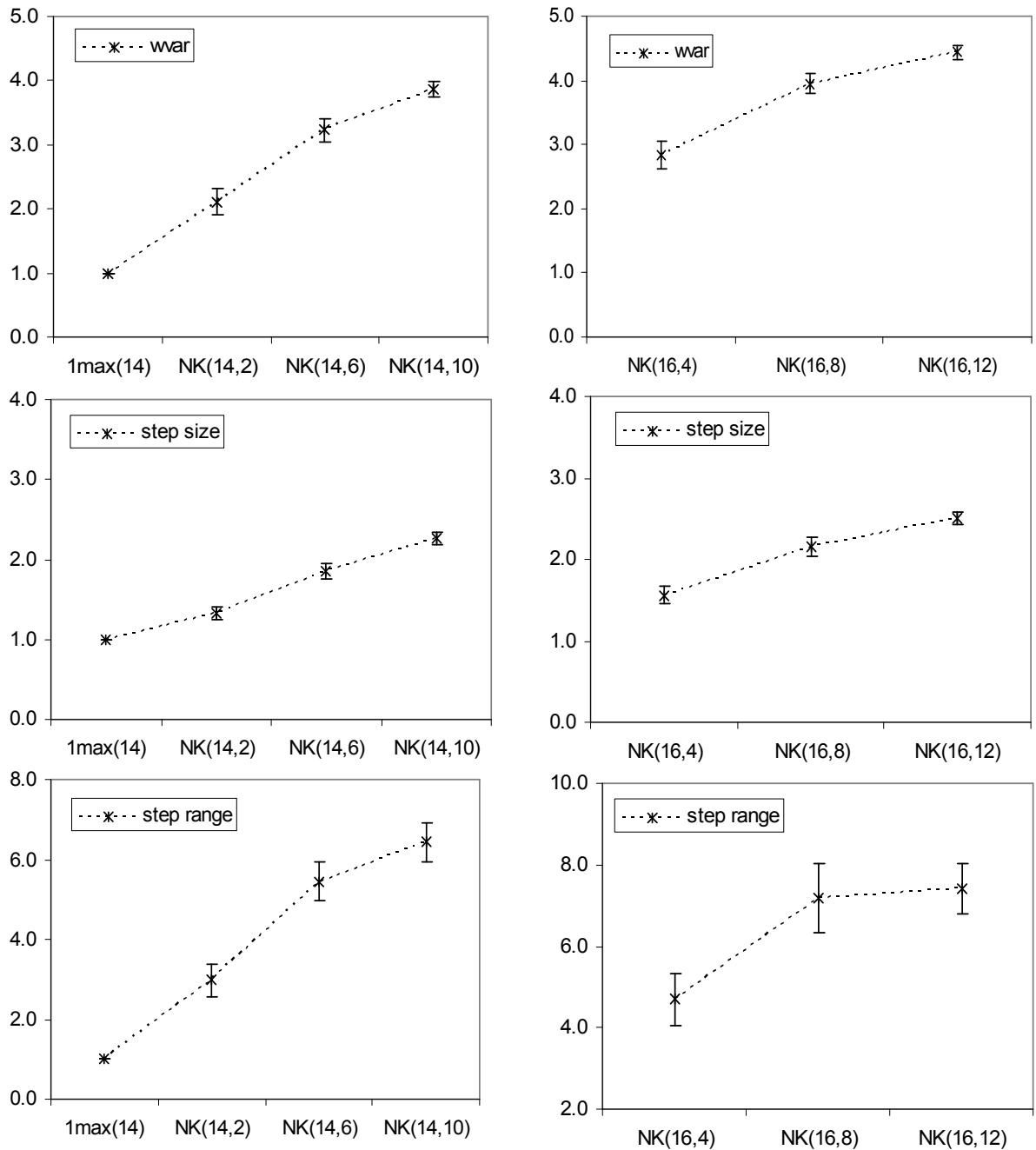

**Fig. 2** Top to bottom: walk variability, step size and step size range averaged over 30 instances per NK problem. Step size variability (*wvar*) is the number of unique step sizes taken in a walk. Step size range is the difference between the maximum and minimum step size taken in a walk. These measurements are made for each walk, then summarized over all walks per NK instance, and then summarized again over all NK instances per NK problem. Step size is the average step size when the step sizes of all walks per problem instance are considered. Error bars indicate 95% confidence interval around the final average values. We note that step size range increases with increase in K and appears to reach its limit at N/2.

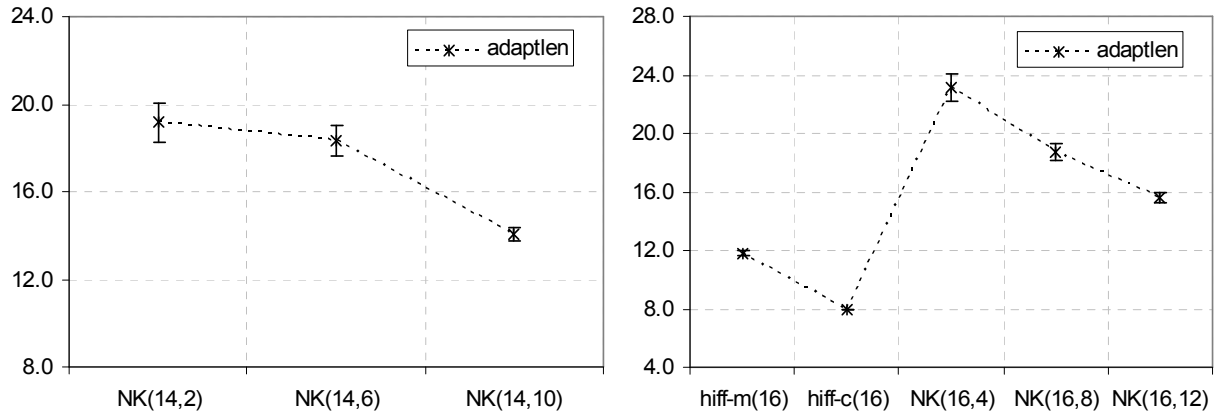

**Fig. 3** Longest sequence of same sized steps in an adaptive walk (adaptive length) per problem instance averaged over 30 instances per NK problem. Error bars indicate 95% confidence interval.

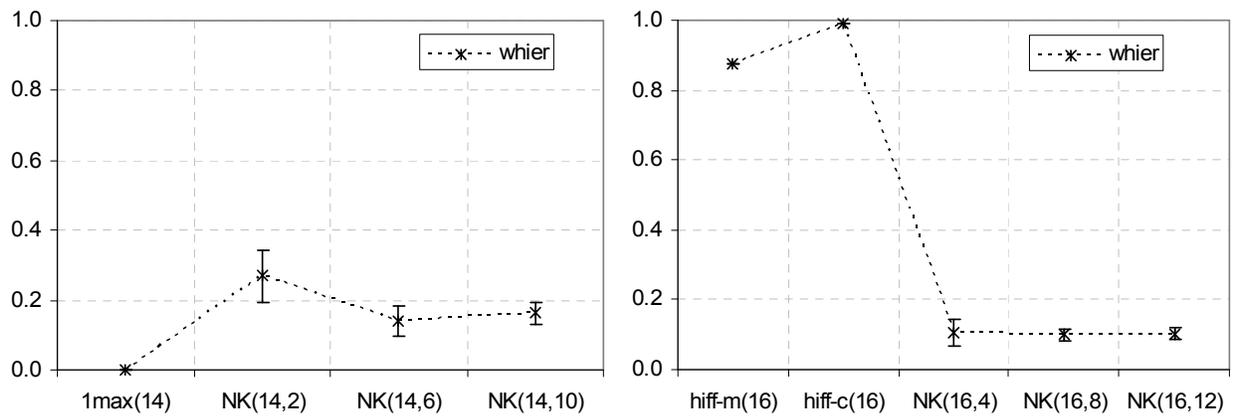

**Fig. 4** Fraction of walks per problem instance which are hierarchical (*whist*) averaged over 30 runs per problem. Although HIFF-C and HIFF-M have non-stochastic fitness functions, the *Limax* algorithm contains a stochastic element and it is possible to generate non-identical walks for the HIFF-C and HIFF-M problems. Error bars indicate 95% confidence interval.

*3.3 Hierarchical walks and anarchic landscapes*

Compressed walks with *cwlen* > 1 with strictly increasing step sizes are hierarchical. *whier* is the fraction of all walks which are hierarchical. Fig. 4 shows *whier* averaged over 30 instances for each problem. For problems with known hierarchical landscapes, i.e. HIFF-C and HIFF-M, *whier* is close to 1.0 as expected. For problems not known to have hierarchical landscapes, i.e. the NK problems, *whier* is closer to 0.0 as expected. *whier* is 0.0 for degenerate cases like One-Max whose walks all compress to a single step size. Knowing *whier* can help narrow down the possibilities for the next move operation and thereby reduce uncertainty in a search. From the perspective of a stochastic search algorithm, a higher *whier* hints that it may not be useful to decrease the current step size. The HIFF and NK fitness landscapes have both been termed rugged. However, *whier* makes a further distinction between the two. There is structure in the ruggedness of the HIFF landscapes in the form of hierarchy, but not so in the NK landscapes particularly with larger Ns and Ks. Thus, the NK landscapes are more anarchic in their ruggedness, and demand more complicatedness from stochastic search algorithms.

## 4. Network analysis

### 4.1 Construction

A directed weighted network with multiple edges is constructed from the set of walks and set of steps from each NK problem (instance). Each node in a *Limax* network represents a search point or unique string configuration. An edge is placed from node *x* to node *y* and labeled *z* if and only if there exists a walk where *Limax* moved from node *x* to node *y* using step size *z*. It should be clear that unlike LONs, the edge weights in a *Limax* network are not transition probabilities between nodes (although we do not rule out the possibility that they may be suitably transformed).

### 4.2 General

Since every *Limax* walk terminates at a global optimum, a *Limax* network forms a single connected component for the single global optimum NK problem. As there is a *Limax* walk commencing from every point in a search space, the number of nodes *V* in a *Limax* network is $2^N$. Nodes without incoming edges are *source nodes*; nodes without outgoing edges are *sink nodes*. For all *Limax* networks for the NK problems, there is only one sink node, which represents the global optimum. For a given N, as K increases, the number of unique edges decreases indicating less expansive node visitation pattern; and the number of source nodes increases (Fig. 5). This is related to walks (*wlen*) getting shorter as K gets larger (Fig. 1).

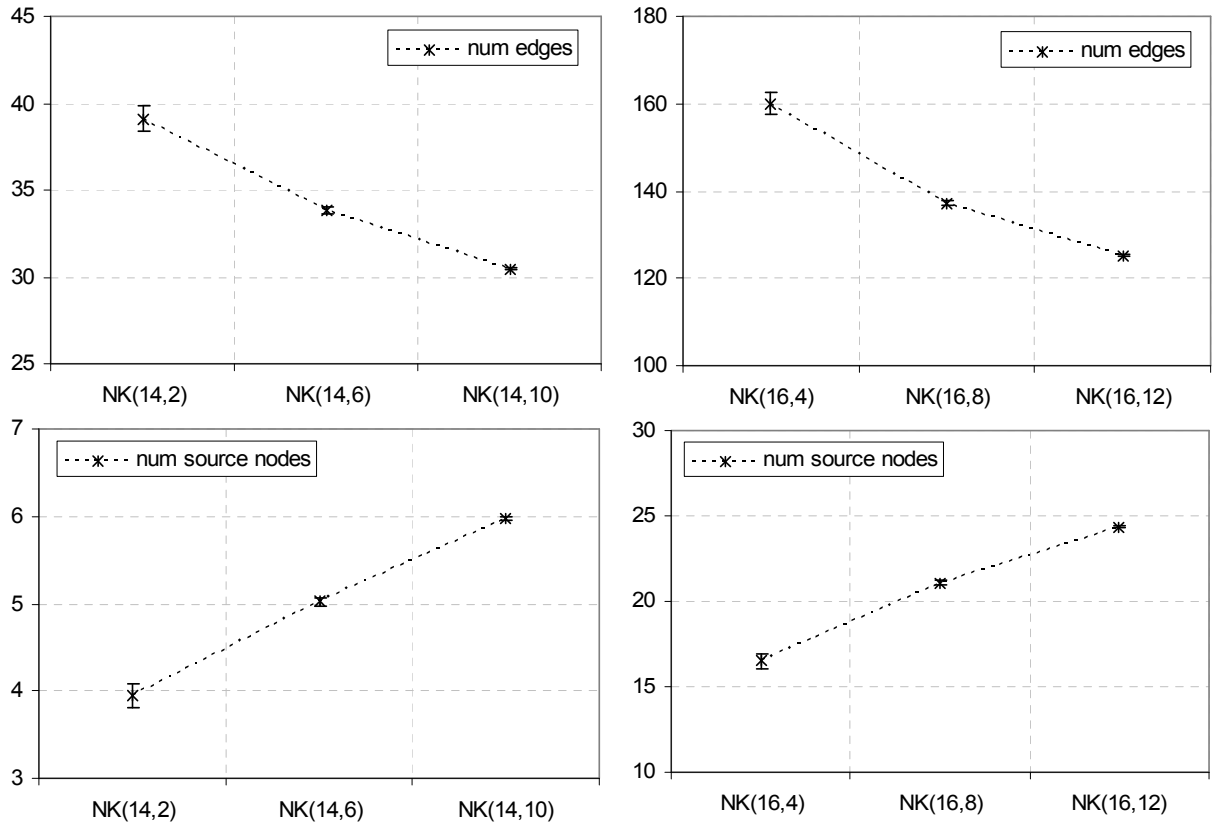

**Fig. 5** Number of edges (multiple edges between nodes are counted as 1), and number of source nodes per *Limax* network averaged over 30 instances per NK problem. All y-axes are in thousands. Error bars indicate 95% confidence interval.

*4.3 Node Degree and Strength*

A node's in-degree is the number of edges ending in it. A node's out-degree is the number of edges starting from it. The sum of a node's in- and out- degrees gives the node's degree. Due to the infinite reach of *Limax*, the search space has no effective local optima, only global optima (one global optimum in the case of NK problems). As with Doye's [2002] *inherent networks*, the *Limax* networks also have scale-free degree distributions (Table 1, Fig. 6). However, *Limax* networks are different from inherent networks, which form the basis of LONs.

[Barthelemy et al 2005] defined the concept of node strength as counterpart to node degree for weighted networks. A node's strength is the sum of the weights of its edges. For *Limax* networks, this is the sum of the step sizes associated with the edges adjacent to a node. A node's in- and out- strengths are defined accordingly. For reasons that will become clearer later (section 4.4), we define a second kind of node strength; one that sums the inverse of the weights of the edges of a node. We call the latter *invstep-strength*, and the former *step-strength*. Both step-strength distribution and invstep-strength distribution of *Limax* networks are scale-free (Table 1, Figs. 7 & 8).

**Table 1** Average (top), standard deviation (middle) and median (bottom) values of two of the 30 *Limax* networks per NK problem chosen at random.

| Problem | In degree | Out degree | Degree | In step strength | Out step strength | Step strength | In invstep strength | Out invstep strength | Invstep strength |
|---|---|---|---|---|---|---|---|---|---|
| NK (14, 2) | 8.44 / 161.71 / 2 | 8.44 / 98.92 / 3 | 16.88 / 235.56 / 5 | 13.59 / 485.24 / 2 | 13.59 / 430.06 / 3 | 27.19 / 833.41 / 5 | 7.10 / 98.97 / 2 | 7.10 / 42.45 / 3 | 14.20 / 124.40 / 5 |
| | 10.43 / 168.90 / 2 | 10.43 / 110.29 / 3 | 20.85 / 254.94 / 5 | 10.57 / 183.53 / 2 | 10.57 / 114.97 / 3 | 21.15 / 268.16 / 5 | 10.35 / 161.90 / 2 | 10.35 / 109.09 / 3 | 20.71 / 249.14 / 5 |
| NK (14, 6) | 11.66 / 221.29 / 1 | 11.66 / 180.57 / 2 | 23.32 / 383.10 / 3 | 22.77 / 970.97 / 1 | 22.77 / 910.15 / 2 | 45.54 / 1461.47 / 3 | 9.21 / 143.67 / 1 | 9.21 / 116.82 / 2 | 18.42 / 225.30 / 3 |
| | 9.04 / 154.73 / 1 | 9.04 / 87.03 / 2 | 18.09 / 215.98 / 3 | 16.32 / 622.05 / 1 | 16.32 / 422.53 / 2 | 32.64 / 889.81 / 3 | 7.31 / 66.17 / 1 | 7.31 / 37.62 / 2 | 14.62 / 95.31 / 3 |
| NK (14, 10) | 10.89 / 208.38 / 1 | 10.89 / 164.50 / 2 | 21.78 / 352.96 / 3 | 19.98 / 487.26 / 1 | 19.98 / 407.45 / 2 | 39.97 / 845.52 / 3 | 8.03 / 156.81 / 1 | 8.03 / 124.85 / 2 | 16.07 / 244.54 / 3 |
| | 8.84 / 155.30 / 1 | 8.84 / 88.04 / 2 | 17.69 / 217.61 / 3 | 19.65 / 793.47 / 1 | 19.65 / 535.25 / 2 | 39.29 / 1091.02 / 3 | 6.18 / 62.30 / 1 | 6.18 / 35.08 / 2 | 12.37 / 86.82 / 3 |
| NK (16, 4) | 13.04 / 375.87 / 2 | 13.04 / 275.26 / 3 | 26.08 / 607.10 / 5 | 17.03 / 531.28 / 2 | 17.03 / 524.59 / 3 | 34.06 / 952.82 / 5 | 11.86 / 349.15 / 2 | 11.86 / 248.63 / 3 | 23.72 / 541.69 / 5 |
| | 10.39 / 342.04 / 2 | 10.39 / 226.89 / 3 | 20.79 / 520.97 / 5 | 22.20 / 1733.92 / 2 | 22.20 / 1332.64 / 3 | 44.41 / 2757.67 / 5 | 7.95 / 126.19 / 2 | 7.95 / 65.26 / 3 | 15.90 / 171.08 / 5 |
| NK (16, 8) | 12.08 / 367.03 / 1 | 12.08 / 263.05 / 2 | 24.16 / 585.04 / 3 | 24.05 / 1079.53 / 1 | 24.04 / 1030.78 / 2 | 48.09 / 1888.36 / 3 | 9.03 / 218.98 / 1 | 9.03 / 147.01 / 2 | 18.06 / 320.80 / 3 |
| | 10.28 / 316.62 / 1 | 10.28 / 186.37 / 2 | 20.56 / 452.15 / 3 | 23.63 / 1641.21 / 1 | 23.63 / 1028.55 / 2 | 47.25 / 2241.92 / 3 | 7.27 / 113.01 / 1 | 7.27 / 69.04 / 2 | 14.55 / 160.50 / 3 |
| NK (16, 12) | 9.93 / 320.08 / 1 | 9.93 / 192.19 / 2 | 19.86 / 461.78 / 3 | 25.68 / 1973.65 / 1 | 25.68 / 1286.88 / 2 | 51.37 / 2776.22 / 3 | 6.40 / 82.41 / 1 | 6.40 / 48.16 / 2 | 12.80 / 121.78 / 3 |
| | 11.89 / 432.53 / 1 | 11.89 / 348.67 / 2 | 23.77 / 742.82 / 3 | 30.03 / 1758.13 / 1 | 30.03 / 1586.57 / 2 | 60.05 / 3051.31 / 3 | 7.13 / 145.29 / 1 | 7.13 / 102.83 / 2 | 14.27 / 235.25 / 3 |

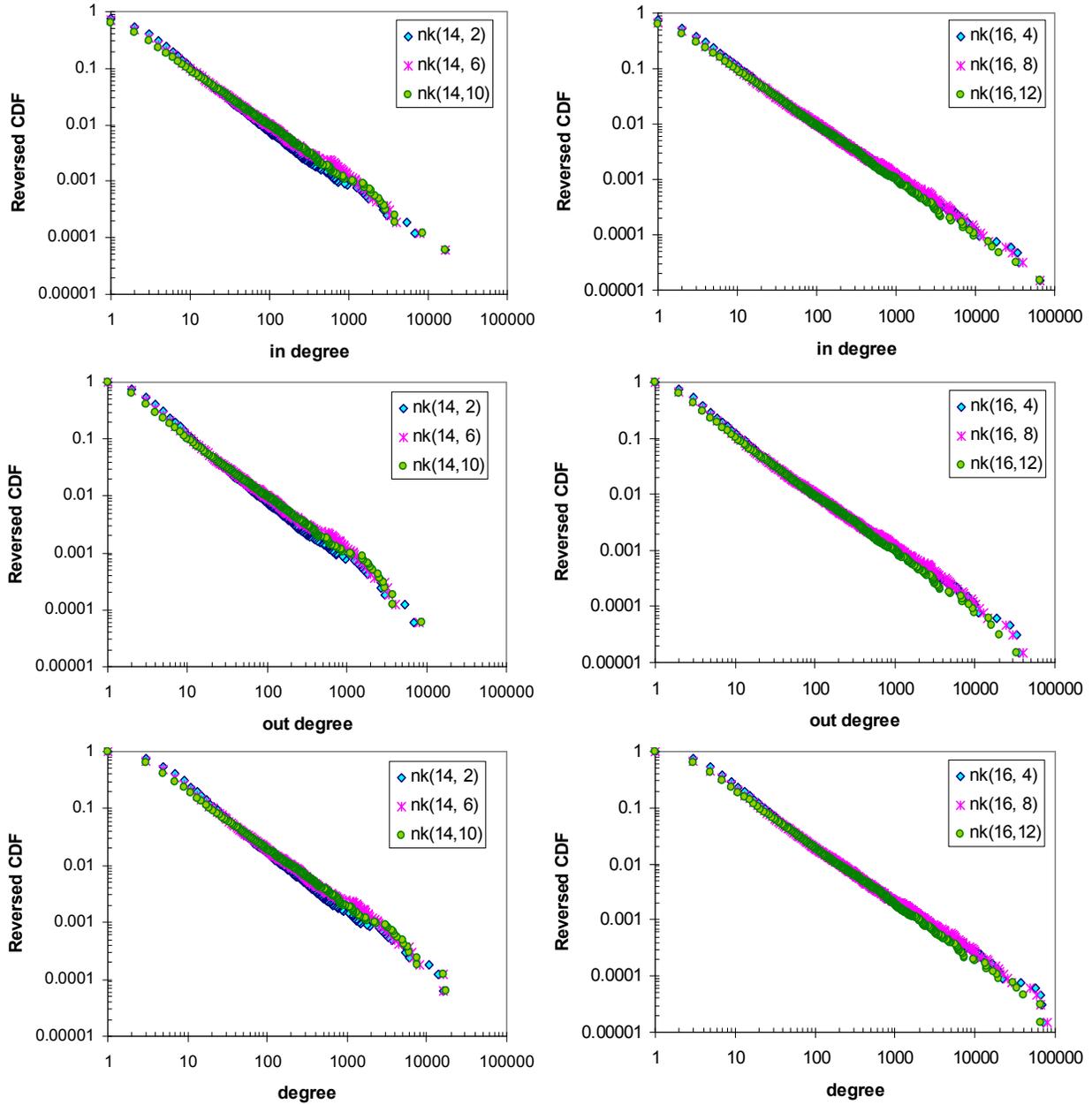

**Fig. 6** Reversed cumulative degree distribution of one *Limax* network chosen at random per NK problem on a double log plot.

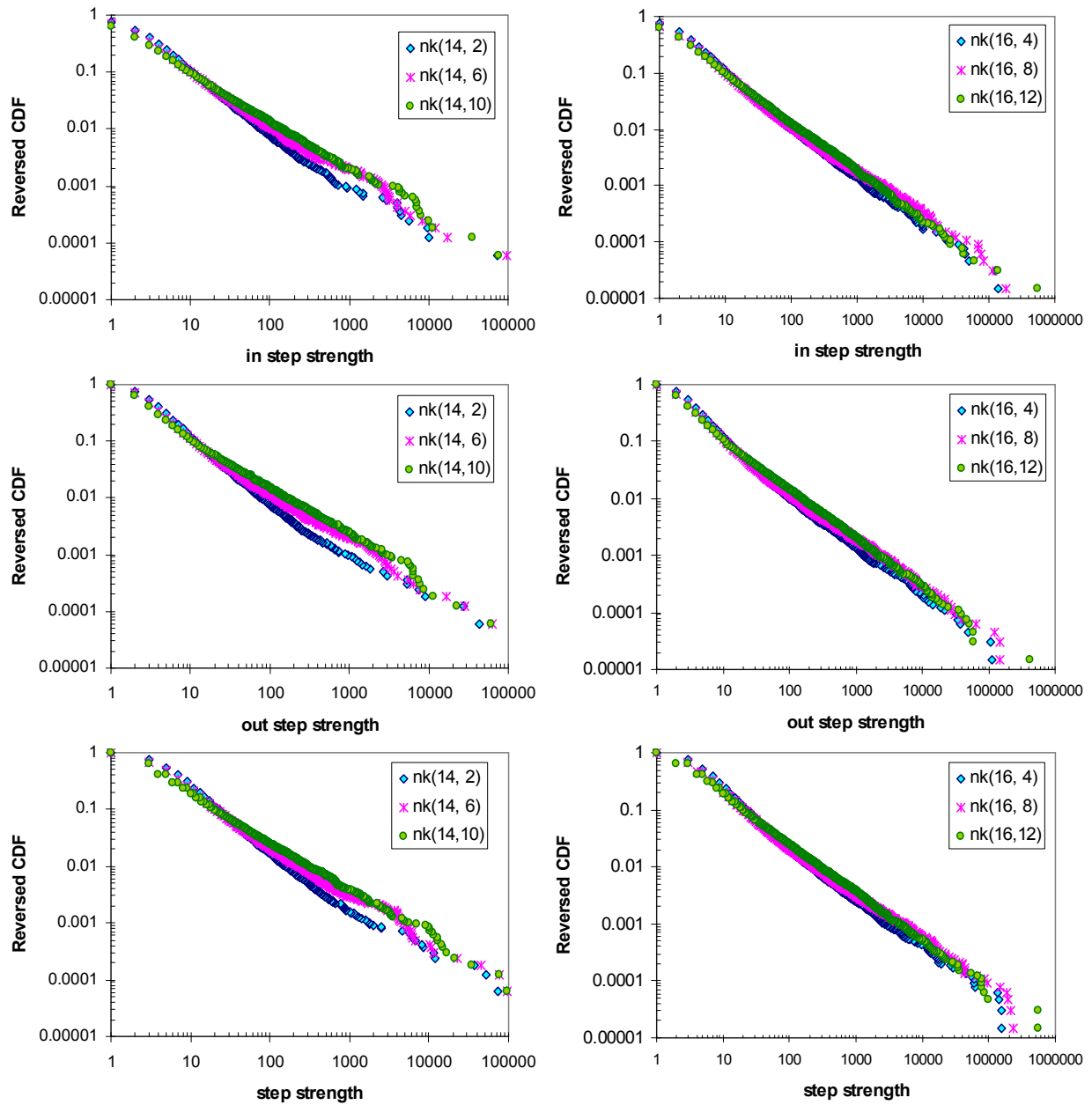

**Fig. 7** Reversed cumulative node step-strength distribution of one *Limax* network chosen at random per NK problem on a double log plot.

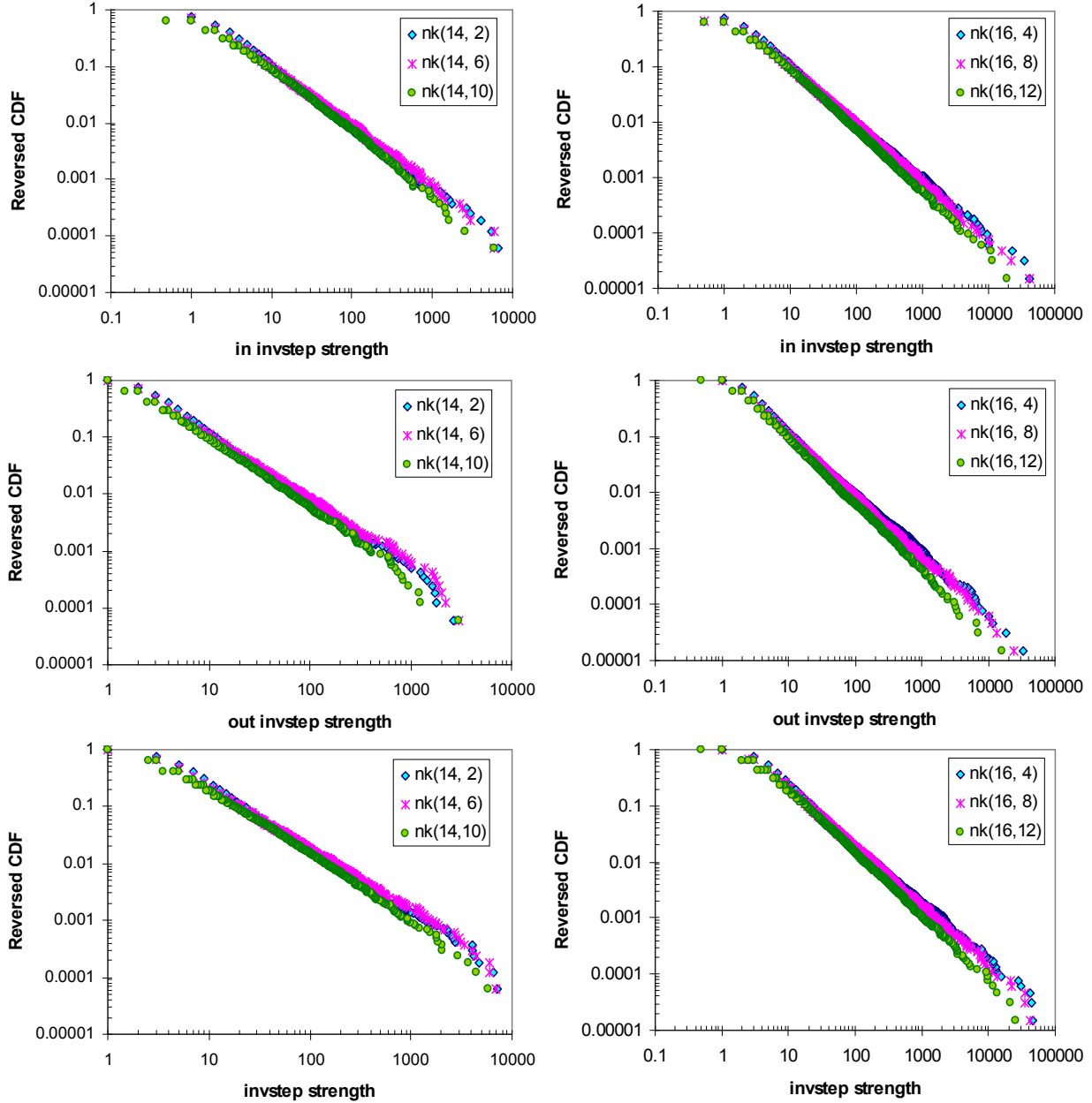

**Fig. 8** Reversed cumulative node invstep-strength distribution of one *Limax* network chosen at random per NK problem on a double log plot.

*4.4 Node Viscidity and Local Optimum Potential*

Node *viscidity* is a node's *in-invstep-strength / out-invstep-strength* when *out-invstep-strength* > 0, and *in-invstep-strength* when *out-invstep-strength* is 0 (Table 3). A node's *in-invstep-strength* is the sum of the inverse of the weights (step size) on its incoming edges. A node's *out-invstep-strength* is the sum of the inverse of the weights (step size) on its outgoing edges. A node's viscidity reflects its ability to pull walks towards it and trap them at it. A node is more likely to attract walks to it if it is easy to reach (close by many nodes so that small step sizes are required to move to it), and it is profitable to do so (it has high fitness or is relatively more fit than many surrounding nodes). In a departure from many landscape studies, we leave the notion of a node's neighbourhood vague to accommodate the *Limax* search algorithm which has a dynamic search radius. A node is more likely to trap walks if a large step is required to leave or push away from it. Hence, we postulate that nodes with larger viscidity values are more likely to be local optimum nodes, and we use a node's viscidity to quantify its local optimum potential. Using the inverse of step sizes creates a positive relationship between the likelihood of entering or leaving a node and the weight of edges (step size) adjacent to the node. Larger step sizes (smaller inverse step sizes) make it less likely or more difficult for a walk to enter or leave a node.

To evaluate the ability of node viscidity (pull using *invstep-strength*) to identify local optima, we need some way to identify local optima independently. We do this in two ways. First, following the more common method, we compute *plf*, the fraction of less fit strings in the 1-bit flip neighbourhood of a string, for each string (node). By definition, a local optimum is fitter than all its neighbours. Thus strings with *plf* = 1.0 are marked as local optima. In the second method, information in a *Limax* network is used to calculate *los*, a local optimum score for each node, as outlined in Fig. 9. Nodes with larger *los* values are interpreted as more likely local optima candidates. The *los* method allows for flexible node neighbourhoods.

```
if (in_max_stepsize == 0)
    score = 0.0
else
    if (out_min_stepsize == 0)
        score = 7.0
    else
        score = 0.0
        if (out_mode_stepsize > in_mode_stepsize)
            score += (out_mode_stepsize - in_mode_stepsize)
        if (out_avg_stepsize > in_avg_stepsize)
            score += (out_avg_stepsize - in_avg_stepsize)
        if (out_min_stepsize > in_max_stepsize)
            score += (out_min_stepsize - in_max_stepsize)
```

**Fig. 9** Algorithm to calculate *los*, local optimum score for each node. `in_max, in_avg` and `in_mode` respectively are the maximum, average and most frequently occurring step size in the set of incoming edges for a node. `out_min, out_avg` and `out_mode` respectively are the minimum, average and most frequently occurring step size in the set of outgoing edges for a node. Examples are available in Table 4. The value 7.0 does not matter in the context of this work since the global optimum is identified independently of *los*, and treated specially (mostly the global optimum and source nodes are excluded from network analysis due to their extreme pull- values).

Both the *plf* and *los* methods identified very similar number of local optima for each problem, and as expected, the number of local optima identified increases as K increases for a given N (Table 2). The *plf* method identified slightly more local optima than the *los* method. This difference is understandable given the more restricted neighbourhood of the *plf* method; but it does not necessarily make *plf* a better standard to measure against than *los*. While all local optima identified by the *los* method were almost always identified by the *plf* method (Table 2 column 4), the same cannot be said about the converse and the discrepancy increases with problem difficulty (Table 2 column 5). On average, the *los* method missed 14 local optima for NK(16, 8) and 55 local optima for NK(16, 12).

**Table 2** Comparison of local optima statistics identified by *plf* and *los* methods.

| NK | Median, Avg, Std. dev. Number of *plf* local optima | Median, Avg, Std. dev. Number of *los* local optima | Median, Avg, Std. dev. *plf* for *los*\* | Median, Avg, Std. dev. *plf* – *los*# |
|---|---|---|---|---|
| (14,2) | 17.00, 19.33, 14.43 | 16.50, 19.03, 14.08 | 1.00, 1.00, 0.00 | 0.00, 0.30, 0.65 |
| (14,6) | 145.00, 142.77, 24.39 | 144.00, 139.60, 23.59 | 1.00, 1.00, 0.00 | 3.00, 3.17, 1.93 |
| (14,10) | 482.00, 479.03, 30.88 | 466.00, 464.70, 30.32 | 1.00, 1.00, 0.00 | 15.00, 14.33, 3.49 |
| (16,4) | 98.50, 107.53, 53.06 | 97.00, 104.97, 51.16 | 1.00, 1.00, 0.00 | 2.00, 2.60, 2.86 |
| (16,8) | 659.00, 656.13, 80.58 | 639.50, 642.17, 78.24 | 1.00, 1.00, 0.00 | 14.50, 14.00, 3.99 |
| (16,12) | 1858.50, 1842.70, 73.85 | 1803.00, 1787.50, 70.18 | 1.00, 1.00, 0.00 | 55.00, 55.27, 6.89 |

\* *plf* score of local optima identified by the *los* method averaged over 30 instances of each NK problem. An average of 1.00 tells us that nodes with *los* score > 0.0 have *plf* = 1.0, i.e. they are fitter than their 1-bit flip neighbours.
# Pair-wise difference between the number of *plf* local optima and the number of *los* local optima averaged over 30 instances of each NK problem.

We compare the ability of node viscidity (pull using *invstep-strength*) to identify local optima against two other possible measures: (i) pull using degree (in-degree / out-degree); and (ii) pull using *step-strength* (in-step-strength / out-step-strength). Values are assigned to the three pull measures in a similar way (Table 3). If a node's in-degree, in-step-strength or in-invstep-strength is 0 (a node with 0 in-degree will also have 0 in-step-strength and 0 in-invstep-strength; the same goes for corresponding out-values), the node cannot be a local optimum (since it did not attract any *Limax* walks to it) and its pull value for the three measures is 0. If a node has a positive (> 0) in-degree, in-step-strength or in-invstep-strength (a node with positive in-degree will also have positive in-step-strength and positive in-invstep-strength; the same applies to corresponding out- values) and a 0 out-degree, out-strength or out-invstep-strength, then it most definitely is a local optimum (it is actually the global optimum in the *Limax* networks under study), and its pull value for the three measures is the corresponding in- value. If both in- and out- degree, step-strength or invstep-strength of a node are positive, then the node maybe a local optimum and its pull value for the three measures is the ratio of in- to out- values. A node's local optimum potential increases with increases in pull using degree, and with increases in node viscidity (pull using invstep-strength). However, excluding nodes with 0 out- values, a node's local optimum potential decreases with increases in pull using step-strength.

**Table 3** Pull value assignment

| in- | out- | LO? | pull |
|---|---|---|---|
| 0 | 0 | No | 0 |
| 0 | + | No | 0 |
| + | 0 | Yes | in- |
| + | + | Maybe | in- /out- |

**Table 4** An example: actual values from run #29 of NK(14, 10)

| #29 | | | Step size | | | Pull- | | |
|---|---|---|---|---|---|---|---|---|
| nid | *plf* | *los* | out_min in_max | out_avg in_avg | out_mode in_mode | degree | step-strength | invstep-strength |
| 17404 | 1 | 3.6192 | 4<br>4 | 4<br>2.3808 | 4<br>2 | 5055 / 5056<br>= 0.9998 | 12035 / 20224<br>= 0.5951 | 2401.17 / 1264<br>= 1.8997 |
| 29681 | 1 | 1.4984 | 2<br>3 | 2<br>1.5016 | 2<br>1 | 313 / 314<br>= 0.9968 | 470 / 628<br>= 0.7484 | 246.83 / 157<br>= 1.5722 |
| 36743 | 1 | 6 | 3<br>1 | 3<br>1 | 3<br>1 | 195 / 196<br>= 0.9949 | 195 / 588<br>= 0.3316 | 195 / 65.33<br>= 2.9847 |

**Table 5** Median, average and standard deviation of pull values for two of the 30 *Limax* networks chosen at random. Given the large standard deviations relative to the mean, all three pull measures are not normally distributed. Remarkably they have very similar median values.

| NK Problem | Pull-degree median, avg, std. dev. | Pull-step-strength median, avg, sd | Pull-invstep-strength median, avg, sd |
|---|---|---|---|
| (14, 2) | 0.67, 1.52, 127.99 | 0.67, 2.54, 258.47 | 0.67, 1.18, 84.49 |
|  | 0.67, 1.56, 127.99 | 0.67, 1.71, 146.77 | 0.67, 1.49, 118.60 |
| (14, 6) | 0.50, 1.50, 127.99 | 0.50, 7.00, 832.24 | 0.50, 0.72, 27.38 |
|  | 0.50, 1.50, 127.99 | 0.50, 4.83, 553.11 | 0.50, 0.82, 39.81 |
| (14, 10) | 0.50, 1.44, 127.99 | 0.50, 6.25, 743.06 | 0.50, 0.80, 43.18 |
|  | 0.50, 1.45, 127.99 | 0.50, 1.66, 154.39 | 0.50, 1.40, 119.72 |
| (16, 4) | 0.67, 1.54, 255.99 | 0.67, 1.54, 255.99 | 0.67, 1.54, 255.99 |
|  | 0.67, 1.51, 255.99 | 0.50, 5.67, 1318.92 | 0.67, 0.85, 86.93 |
| (16, 8) | 0.50, 1.48, 255.99 | 0.50, 3.26, 711.60 | 0.50, 1.13, 164.68 |
|  | 0.50, 1.47, 255.99 | 0.50, 6.31, 1492.71 | 0.50, 0.80, 80.96 |
| (16, 12) | 0.50, 1.44, 255.99 | 0.50, 7.40, 1780.83 | 0.50, 0.61, 39.71 |
|  | 0.50, 1.44, 255.99 | 0.50, 5.27, 1236.80 | 0.50, 0.68, 56.38 |

To evaluate the ability of the three pull measures to identify local optima, we first filter out the global optimum node and nodes with 0 pull values, and sort the remaining nodes in order of decreasing local optimum potential, i.e. in descending order of pull-degree values, in ascending order of pull-step-strength values and in descending order of node viscidity or pull-invstep-strength values. Next, the corresponding *plf* and *los* values for the sequence of nodes are mapped, and the resultant *plf* and *los* sequences are analyzed for *false positives*, *edit distance* and *rank distance* (the last two only applies to *los*).

An ideal pull measure would identify all local optima and only local optima, and where applicable, rank the nodes according to their local optimum potential. Thus, an ideal node sequence is one with all local optimum nodes placed consecutively starting from the beginning of the sequence. In terms of *plf* and *los* sequences, this means all positive values appear before any 0 values (non-local optimum nodes have *plf* = 0.0 or *los* = 0.0). Let *x* be the position of the last positive value in a *plf* or *los* sequence. The number of 0's interspersed between the start of a *plf* or *los* sequence and *x* is the number of false positives for a pull measure. The *error rate* for a pull measure is the number of false positives divided by $V (= 2^N)$, the number of nodes in a *Limax* network.

The *plf* method characterizes nodes as either local optima or non-local optima; thus its sequences comprise 1's and 0's only. With the *los* method, there are gradations of "local optimum-ness", where larger values reflect stronger local optimum potential. Define *loseq* as the *los* values sorted in descending order; and *loseq-x* as a *los* sequence generated by a pull method with positions after *x* (defined above) chopped off and all false positives removed. Edit distance is the number of pairwise mismatch values between *loseq* and *loseq-x*. Rank distance is the sum of the pairwise absolute difference between *loseq* and *loseq-x*. It reflects how differently a pull method ranks nodes by local optimum potential from the so-called ideal ranking constructed by *los*. The pull measure with the fewest false positives and the smallest edit and rank distances, is considered to contain information which best reflects local optimum potential. We find that these conditions are best fulfilled by pull-invstep-strength or node viscidity (Tables 6a & 6b, Figs. 10 & 11).

**Table 6a** Number of instances per NK problem with 0 error rate

| NK | *plf* method | | | *los* method | | |
|---|---|---|---|---|---|---|
| | Pull-degree | Pull-step-strength | Pull-invstep-strength | Pull-degree | Pull-step-strength | Pull-invstep-strength |
| (14,2) | 1 | 6 | 28 | 1 | 6 | 30 |
| (14,6) | 0 | 0 | 11 | 0 | 0 | 17 |
| (14,10) | 0 | 0 | 1 | 0 | 0 | 0 |
| (16,4) | 0 | 0 | 15 | 0 | 0 | 20 |
| (16,8) | 0 | 0 | 2 | 0 | 0 | 0 |
| (16,12) | 0 | 0 | 0 | 0 | 0 | 0 |

**Table 6b** Number of instances per NK problem with 0 edit distance (the *los* method)

| NK | Pull-degree | Pull-step-strength | Pull-invstep-strength |
|---|---|---|---|
| (14,2) | 6 | 12 | 15 |
| (14,6) | 0 | 0 | 0 |
| (14,10) | 0 | 0 | 0 |
| (16,4) | 0 | 2 | 2 |
| (16,8) | 0 | 0 | 0 |
| (16,12) | 0 | 0 | 0 |

The overall pull- results for both *plf* and *los* methods support the same conclusions. Pull-step-strength produced the highest error rate or most false positives, although these statistics show a decline with increase in problem difficulty (and local optima). Nonetheless, the error rate and false positive statistics of pull-step-strength remain well above those of pull-invstep-strength. Pull-degree produced significantly fewer false positives than pull-step-strength, but this advantage disappears as K becomes larger. Pull-degree yielded significantly larger edit distances and rank distances than the other two pull measures whose edit distance and rank distance statistics, though significantly different from each other, are nonetheless much smaller. Pull-step-strength has a slight advantage over pull-invstep-strength in that it produced significantly smaller edit and rank distances. Hence, pull-degree appears to be the worse measure of local optimum potential, and pull-invstep-strength or node viscidity the best.

We attribute the dismal performance of pull-degree to its disregard of edge weights or step sizes. When step sizes are included in the equation, e.g. in pull-step-strength and in pull-invstep-strength, the edit distances and rank distances shrink significantly. Further investigation is required to understand why pull-step-strength generates so many more false positives than pull-invstep-strength. The example in Table 4 (nid = 29681) provides a clue. If we take 0.5 as the median for all three pull values, which from Table 5 seems a reasonable thing to do, node 29681 will be ranked above nodes with median viscidity values by all three pull measures. However, since local optimum potential increases with pull-degree and with pull-invstep-strength, but decreases with pull-step-strength, node 29681 will appear earlier in the sequences produced by pull-degree and by pull-invstep-strength, but later in the sequence produced by pull-step-strength.

Despite its superior performance as a local optimum detector, node viscidity becomes less accurate when problem difficulty increases. The number of instances with 0 error rate falls sharply (Table 6a), and edit and rank distances show an increasing trend, as K increases for a given N (Figs. 10 & 11). This is an area for further research. One possibility is to consider more than 1 degree of separation when computing pull values.

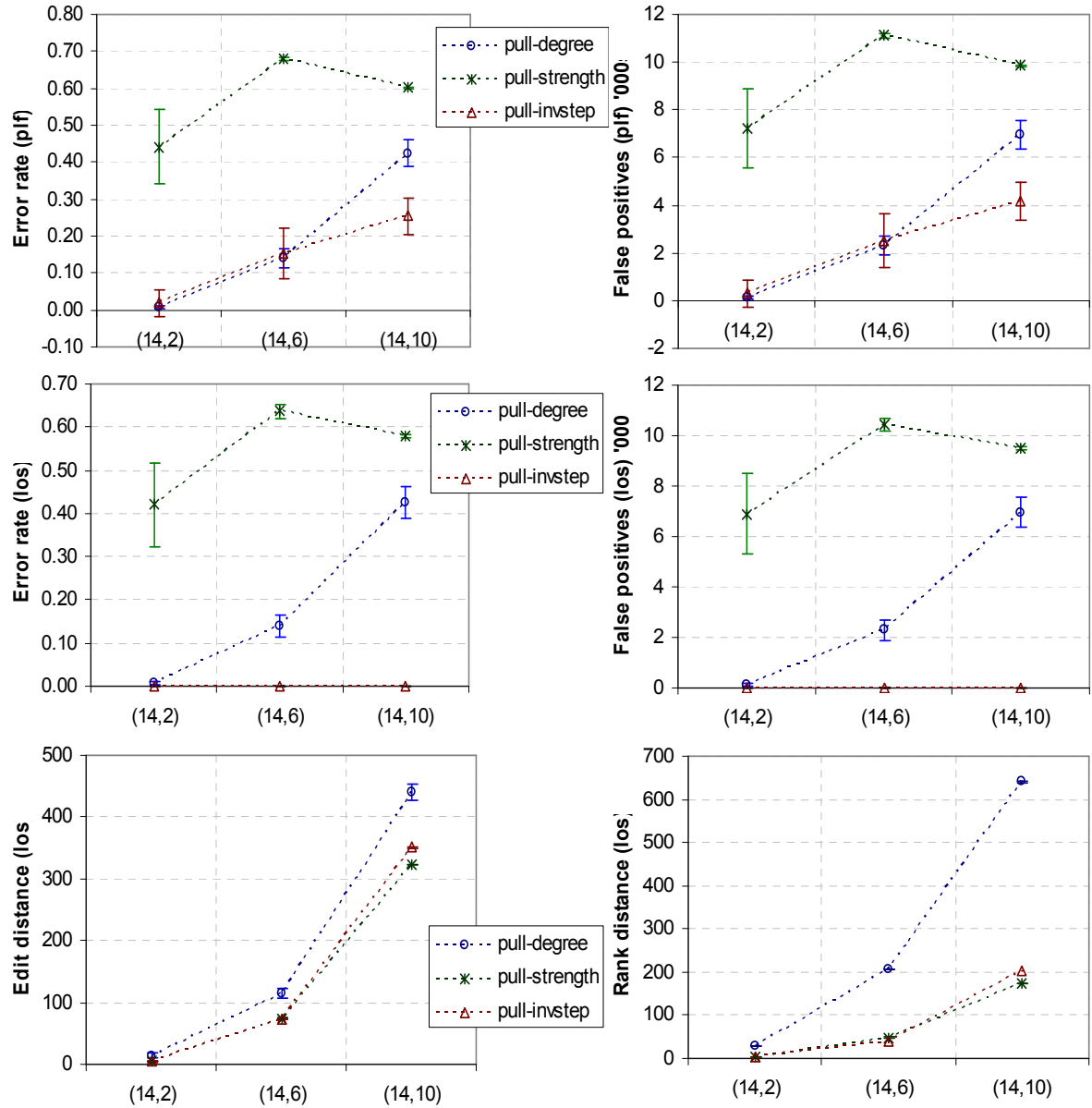

**Fig. 10** Pull- measures evaluated against the *plf* and *los* method of identifying local optima for NK 14 problems. Pull-strength refers to pull-step-strength. Average values over 30 instances per NK problem are reported. The error bars indicate 95% confidence interval. Note the y-axes for false positive plots are in thousands. The same legend applies to all plots.

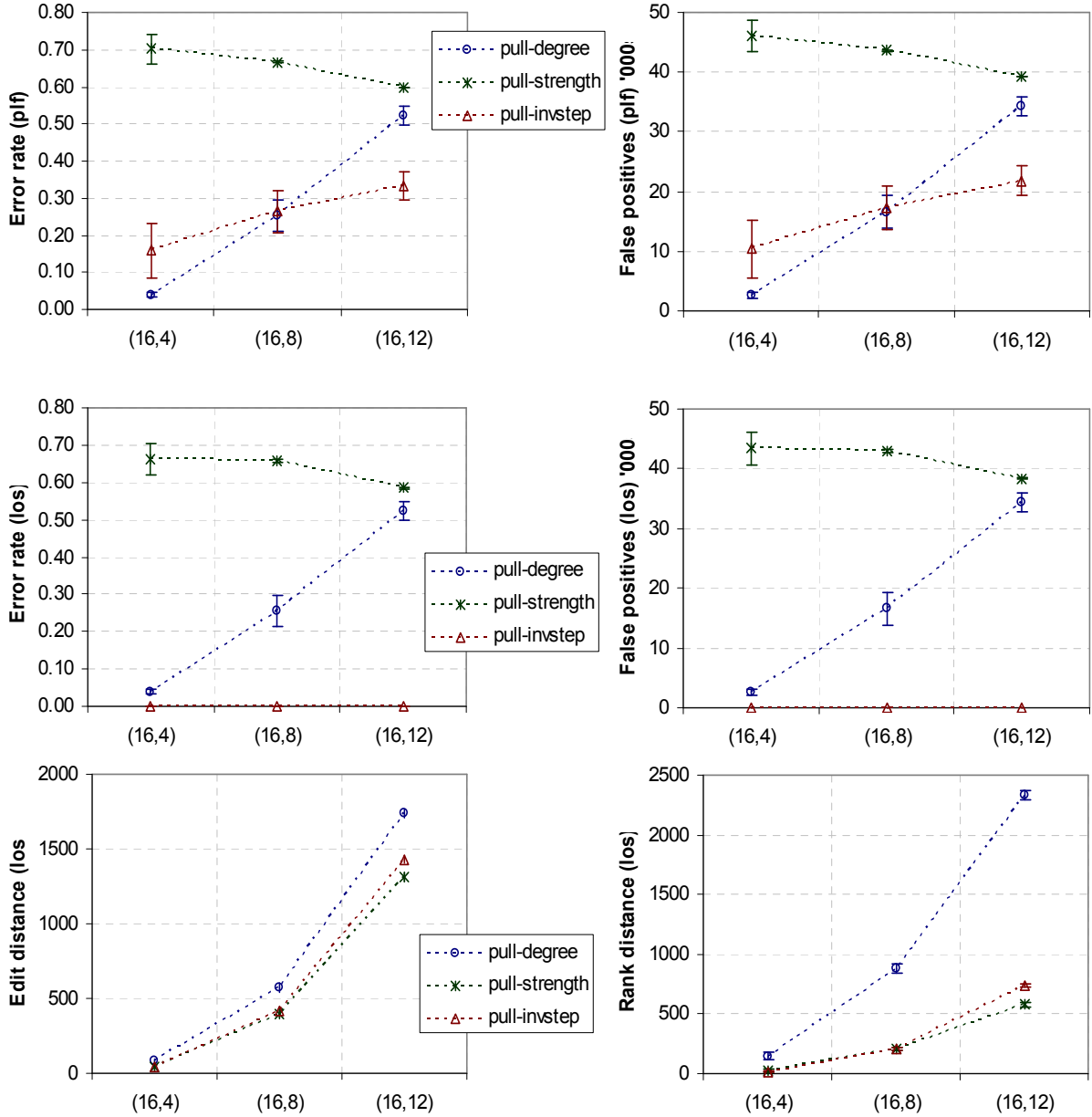

**Fig. 11** Pull- measures evaluated against the *plf* and *los* method of identifying local optima for NK 16 problems. Pull-strength refers to pull-step-strength. Average values over 30 instances per NK problem are reported. The error bars indicate 95% confidence interval. Note the y-axes for false positive plots are in thousands. The same legend applies to all plots.

*4.5 Centrality*

Due to the nature of local optima as points of attraction in a fitness landscape, we expect nodes with high local optimum potential to occupy more central positions in *Limax* networks. Nodes which occupy more central positions in *Limax* networks have high centrality in the sense that many more walks pass through them (global optima and source nodes are excluded by this definition and in the following analyses). To test our hypothesis, we compare the centrality of *top nodes*, nodes with viscosity values in the top quartile, with (i) the centrality of all nodes; and (ii) the centrality of nodes chosen at random to match the size of *top nodes*. Tables 7 and 8 give a feel for the range of values under discussion. From Fig. 12, it is clear that the centrality of *top nodes* is significantly higher than both the centrality of all nodes and the centrality of nodes chosen at random. This confirms the hypothesis that nodes with high local optimum potential occupy more central positions in *Limax* networks. Further, node viscosity and node centrality is strongly positively correlated (Fig. 13). Their respective Spearman's rho values are close to 1.0, although the rho values decrease slightly but significantly (paired t-test pvalue < 0.05) with increases in K given N. However, the relationship between node viscosity and node centrality is non-linear (their respective Pearson's correlation is weak). Nodes with high viscosity are also more central for the HIFF problems (Fig. 12).

**Table 7** Viscosity values at the boundary of the top quartile, and the number of top nodes and top edges averaged over the 30 instances of each NK problem. A top node is one whose viscosity is in the top quartile. A top edge is one whose end points are top nodes.

| NK | Top 25% Viscidity avg, std. dev. | Num top nodes avg, std. dev. | Num top edges avg, std. dev. |
|---|---|---|---|
| (14, 2) | 0.8067, 0.0136 | 4612, 294 | 15154, 1233 |
| (14, 6) | 0.7841, 0.0219 | 4457, 511 | 11699, 1399 |
| (14, 10) | 0.7500, 0.0000 | 4821, 45 | 11335, 150 |
| (16, 4) | 0.8033, 0.0102 | 18897, 888 | 64727, 4633 |
| (16, 8) | 0.7669, 0.0228 | 19353, 2384 | 54170, 6880 |
| (16, 12) | 0.7500, 0.0000 | 19228, 124 | 48572, 499 |

**Table 8** Median, average and standard deviation of centrality values for one of the 30 *Limax* networks chosen at random. Given the large standard deviations relative to the mean, node centrality in any of the three cases, is not-normally distributed.

| | Centrality over All nodes | | | Centrality over Top nodes | | | Centrality over Random nodes | | |
|---|---|---|---|---|---|---|---|---|---|
| NK | median | avg | std. dev. | median | avg | std. dev. | median | avg | std. dev. |
| (14, 2) | 2 | 8.0449 | 114.8248 | 7 | 23.6629 | 208.8175 | 2 | 6.3018 | 44.6715 |
| (14, 6) | 1 | 8.3514 | 97.5155 | 7 | 24.0476 | 120.4865 | 1 | 9.1575 | 97.6367 |
| (14, 10) | 1 | 8.2648 | 101.7460 | 6 | 22.1487 | 161.6981 | 1 | 7.6198 | 77.1636 |
| (16, 4) | 2 | 11.7721 | 217.9451 | 9 | 35.6977 | 359.4067 | 2 | 11.7356 | 224.8032 |
| (16, 8) | 1 | 10.0058 | 191.6944 | 5 | 24.5107 | 304.6448 | 1 | 11.1230 | 188.4229 |
| (16, 12) | 1 | 9.3700 | 193.7662 | 6 | 25.5185 | 311.5091 | 1 | 7.7898 | 91.5826 |

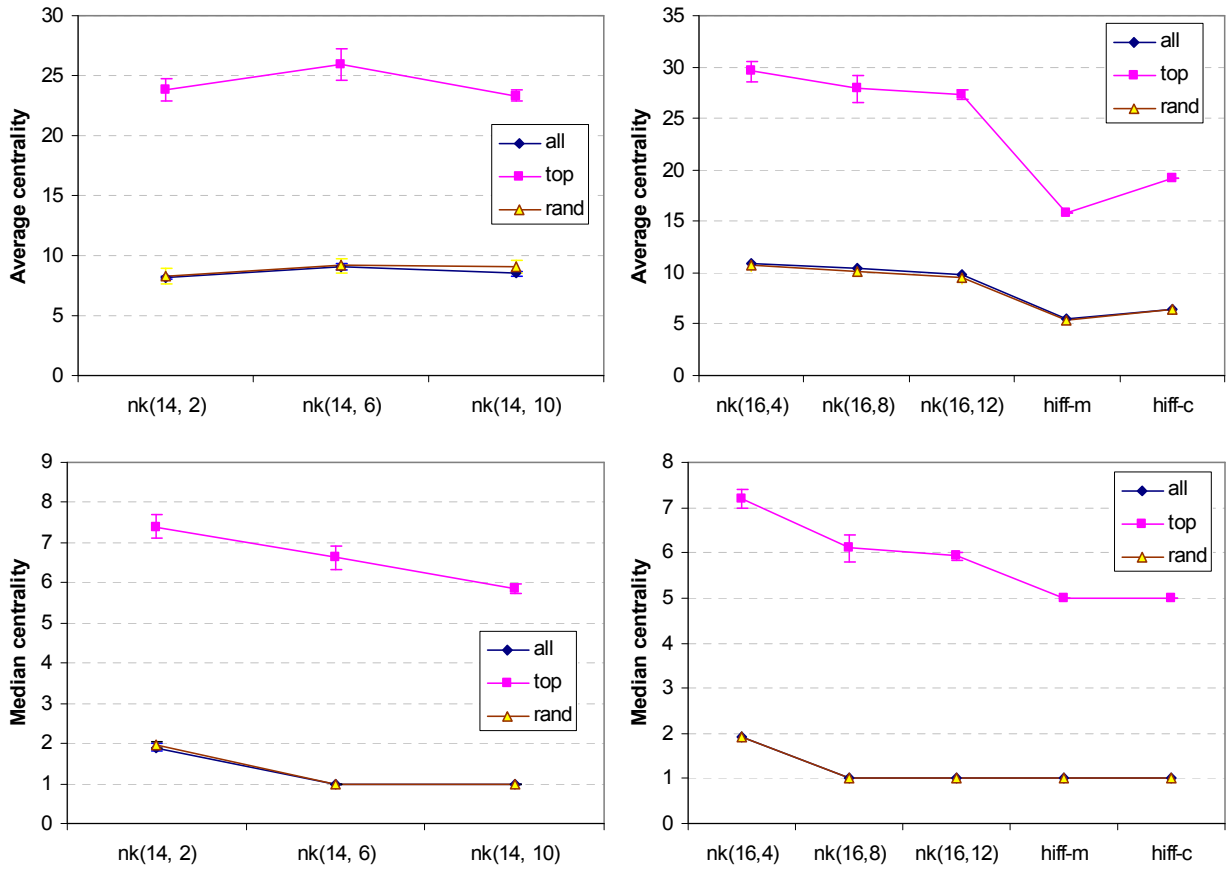

**Fig. 12** Average and median centrality for top nodes, all nodes and random nodes averaged over 30 runs per problem. Error bars indicate 95% confidence interval.

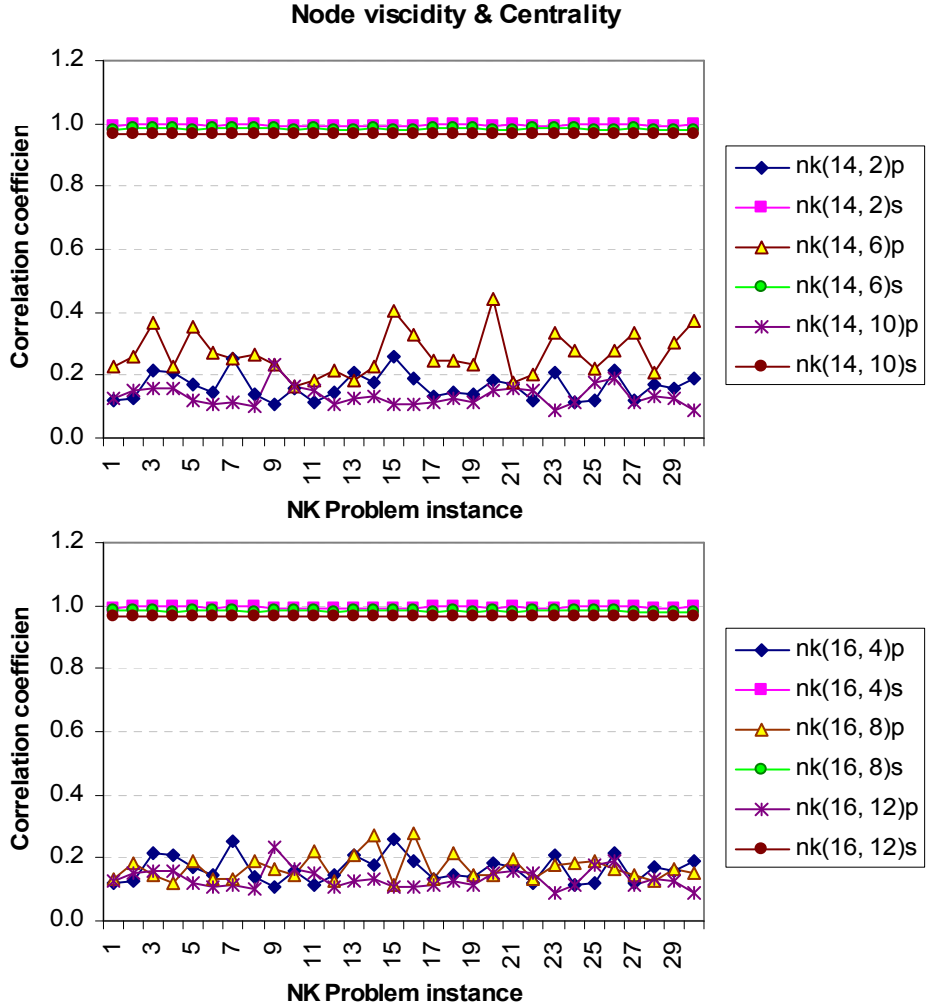

**Fig. 13** Pearson's (p) and Spearman's (s) correlation coefficients for the node viscidity and centrality relationship.

*4.6 Assortativity*

Both the NK and HIFF problems have rugged landscapes, but how are the many local optima distributed in the search space? Are nodes with high local optimum potential more likely to be connected in *Limax* networks or less likely? The effectiveness of a heuristic search algorithm is strongly influenced by the number and distribution pattern of local optima in a search space. Analysis in section 3.3 using hierarchical walks gave some hints. From a network viewpoint, this question can be studied using the notion of assortativity or node attribute mixing. We use the coefficient proposed by [Newman 2002], and the attribute of interest is node viscidity.

We find that the *Limax* networks for the NK problems are disassortative, more so when K is small (Fig. 14 top). This is supported by an increasing double to single edge ratio with increases in K (Fig. 14 bottom). A double edge or top edge (section 4.5) is one where both endpoints have viscosity in the top quartile. A single edge is one where only one endpoint has viscosity in the top quartile.

Results from the node viscosity mixing analysis imply that for more difficult problems (larger K), nodes with high viscosity are more likely to be connected so that a *Limax* walk needs to jump from one local optimum to another, which fits nicely with the idea of a rugged landscape. More extremely, assortativity approaches 0.0 and skirts the positive region in the case of the HIFF problems which is expected given their hierarchical landscapes. The disassortative nature of *Limax* networks for problems with K on the smaller side implies a less bumpy walk towards the global optimum, and the momentum gained by overcoming the distance barrier of an earlier local optimum in a walk may be used to propel it over some distance before encountering another local optima. This is related to adaptive lengths (section 3.2). Such periods of "coasting" or "gliding" are less frequent for problems with K on the larger side. For the HIFF problems, a stochastic search algorithm needs to keep increasing its step size until it reaches a global optimum.

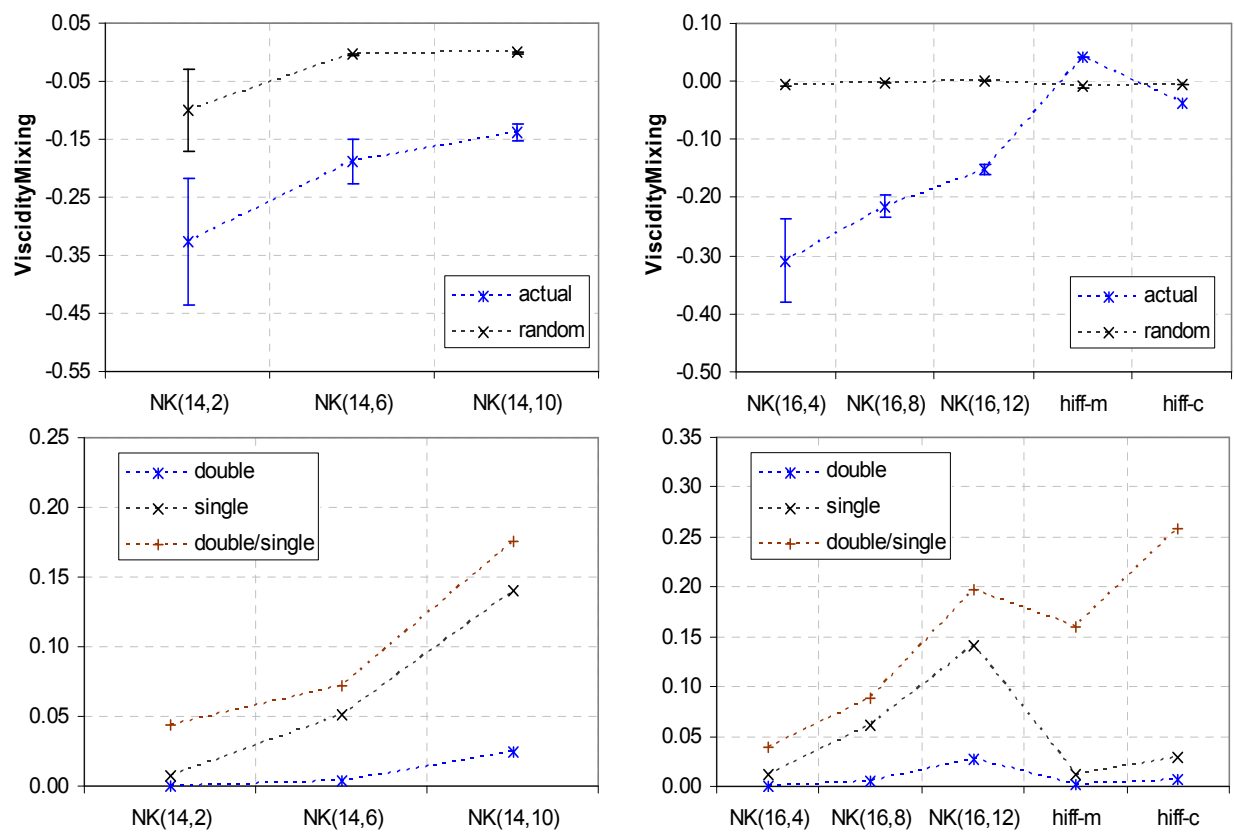

**Fig. 14** Top: assortativity coefficients for *Limax* networks (actual), and for *Limax* networks with node viscosity permuted (random). Bottom: double is the proportion of all edges whose endpoints have top quartile viscosity values; single is the fraction of all edges such that only one endpoint has top quartile viscosity value; double/single is the ratio of double edges to single edges. Reported values are averages over 30 runs per problem. Error bars indicate 95% confidence interval.

A more disassortative *Limax* network when K is small relative to N and a less disassortative *Limax* network when K is larger is not contradictory to the 'Massive Central' phenomenon described in [Kauffman 1993, p. 60]. Fig. 15 shows the average hamming distance between all pairs of nodes with viscidity values above a certain cut-off. Only problems with smaller K have average hamming distances significantly smaller than N/2. This agrees with the known observation that local optima tend to cluster together forming a 'Massive Central' when K is small relative to N, but this effect dissipates as K increases, and the dissipation is faster when the neighbourhood is random.

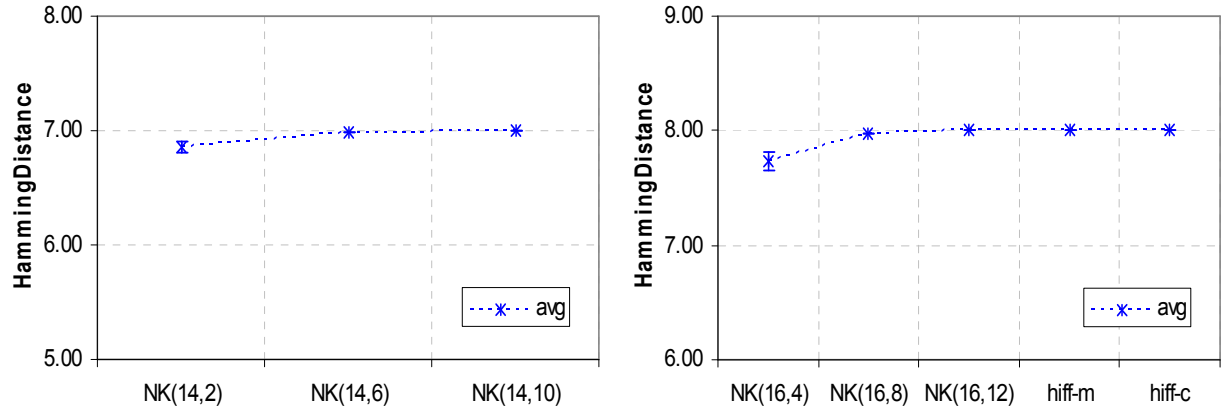

**Fig. 15** Average hamming distance between all pairs of nodes with node viscidity values not less than the top 25% for N=14, and 10% for N=16 (to work within the computation resources available, a higher node viscidity cutoff for N=16 is used to yield fewer qualified nodes). Error bars indicate 95% confidence interval.

## 5. Summary

Slow self-avoiding adaptive walks by an infinite radius search algorithm (*Limax*) are analyzed as themselves, and as the network they form. The study is conducted on several NK problems and two HIFF problems. We find that examination of such "slacker" walks and networks can indicate relative search difficulty within a family of problems, help identify potential local optima, and detect presence of structure in fitness landscapes. The main results are:

(i) Given N, problems with larger K have significantly longer, varied and less compressible walks (section 3.1);

(ii) Given N, the adaptive length (the longest sequence of same sized steps in a *Limax* walk) decreases significantly as K increases (section 3.2);

(iii) The NK landscapes are anarchic compared with the HIFF landscapes where almost all *Limax* walks are hierarchical (section 3.3);

(iv) Several attributes of *Limax* networks have scale-free distributions including degree, strength and invstep-strength (section 4.3);

(v) Node viscosity (sum of the inverse of step sizes on incoming edges to a node divided by the sum of the inverse of step sizes on outgoing edges from the node) can be a measure of local optimum potential (section 4.4);

(vi) Nodes with high viscosity are more centrally located within a *Limax* network (section 4.5);

(vii) Given N, *Limax* networks for smaller K tend to be more disassortative in terms of node viscosity mixing (section 4.6); and

(viii) Nodes with high viscosity tend to clump together in search space when K is small relative to N (section 4.6).

Points (i) – (iii) confirm what is already known about the test problems in terms of their relative search difficulty. Point (iv) suggests a way NK problems can produce scale-free distributions, an issue raised in [Ochoa et al 2008]. Point (v) is the main and we believe novel contribution of this research. Points (vi) – (viii) support point (v) and demonstrate how network analysis based on node viscosity can help illuminate multidimensional search spaces.

**Acknowledgements**

Thanks to Dr. P. Grogono for arranging the use of computer resources at Concordia University, Montreal, Canada.

**References**

Barthelemy, M, Barrat, A., Pastor-Satorras, R., and Vespignani, A. (2005) Characterization and modeling of weighted networks. *Physica A*, 346:34-43.
Doye, J.P.K. (2002) The network topology of a potential energy landscape: A static scale-free network. *Phys. Rev. Lett*. 88, 238701
Doye, J.P.K. and Massen, C.P. (2005) Characterizing the network topology of the energy landscapes of atomic clusters. *Journal of Chemical Physics* 122, 084105
Jones, T. (1995) *Evolutionary algorithms, fitness landscapes and search*. Ph.D. Dissertation, University of New Mexico, New Mexico, USA.
Kauffman, S. A. (1993) *The Origins of Order: Self-organization and Selection in Evolution.* Oxford University Press.
Khor, S. (2009) Exploring the influence of problem structural characteristics on evolutionary algorithm performance. *IEEE Congress on Evolutionary Computation*, pp.3345-3352.
Newman, M.E.J. (2002) Assortativity mixing in networks. *Physical Review Letters* 89, 208701.
Ochoa, G., Tomassini, M., Verel, S. and Darabos, C. (2008) A study of NK landscapes' basins and local optima networks. In Proc. *GECCO* pages 555-562 ACM Press
Verel, S., Ochoa, G. and Tomassini, M. (2008) The connectivity of NK landscapes' basins: A network analysis. In Proc. *Artificial Life* XI, 648-655 MIT Press
Watson, R.A. (2002) *Compositional evolution: Interdisciplinary investigations in evolvability, modularity and symbiosis*. Ph.D. Dissertation, Brandeis University, Massachusetts, USA.